\documentclass[runningheads]{llncs}

\newif\iffinal
\finaltrue

\newif\ifarxiv
\arxivtrue

\usepackage[utf8]{inputenc}
\usepackage[T1]{fontenc}
\usepackage{booktabs}
\usepackage{graphicx}
\usepackage{xspace}
\usepackage[colorlinks=true,allcolors=black]{hyperref}
\usepackage{ifthen}
\usepackage[acronym]{glossaries}
\usepackage[dvipsnames,table]{xcolor}
\usepackage[capitalise]{cleveref} %
\usepackage[sort,compress]{cite}
\usepackage[caption=false,farskip=0pt]{subfig}
\usepackage{arydshln} %

\newcommand\customorcidAuthor[1]{\hspace*{-1mm}
\href{https://orcid.org/#1}{\includegraphics[width=0.3cm]{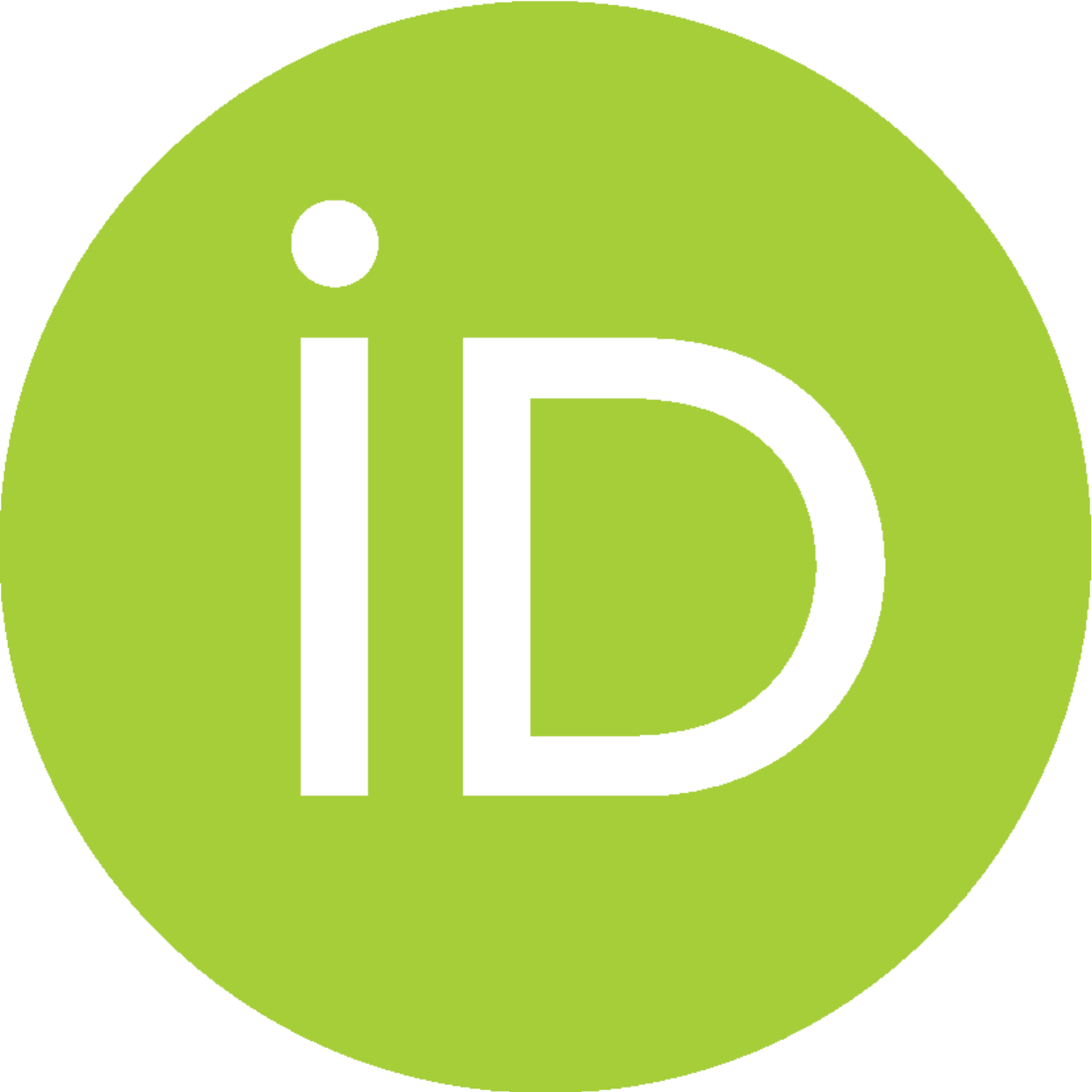}}\hspace*{-1mm}
} %

\newcommand*{\VO}[2][]{\textcolor{blue}{[\textbf{\ifthenelse{\equal{#1}{}}{VO}{VO(#1)}}: #2]}}
\newcommand*{\RL}[2][]{\textcolor{Rhodamine}{[\textbf{\ifthenelse{\equal{#1}{}}{RL}{RL(#1)}}: #2]}}
\newcommand*{\GL}[2][]{\textcolor{brown}{[\textbf{\ifthenelse{\equal{#1}{}}{GL}{GL(#1)}}: #2]}}
\newcommand*{\DM}[2][]{\textcolor{red}{[\textbf{\ifthenelse{\equal{#1}{}}{DM}{DM(#1)}}: #2]}}

\newacronymstyle{long-short-br}
{%
  \GlsUseAcrEntryDispStyle{long-short}%
}%
{%
  \GlsUseAcrStyleDefs{long-short}%
}
\setacronymstyle{long-short-br}

\usepackage{transparent}
\usepackage{tikz}
\ifarxiv
    \newcommand\copyrighttext{%
      \scriptsize This is an author-prepared version. The final published version is available from \textit{Springer} (DOI: \href{https://doi.org/10.1007/978-3-032-39518-4_13}{\textcolor{blue}{https://doi.org/10.1007/978-3-032-39518-4\_13}}).}
    \newcommand\copyrightnotice{%
    \vspace{-5mm}
    \begin{tikzpicture}[remember picture,overlay]
    \node[anchor=south,yshift=80pt,xshift=0pt] at (current page.south) {\fbox{\transparent{0.85}\parbox{\dimexpr0.74\textwidth-\fboxsep-\fboxrule\relax}{\copyrighttext}}};
    \end{tikzpicture}%
    }
\else
\fi

\begin{document}
\title{Revisiting Vehicle Color Recognition in Long-Tailed Surveillance Scenarios}
\titlerunning{Revisiting Vehicle Color Recognition in Long-Tailed Surveillance Scenarios}
\iffinal
    \author{Vinícius~Orrú\inst{1,2}\customorcidAuthor{0009-0001-9478-8710} \and Bruno~H.~Foggiatto\inst{1}\customorcidAuthor{0009-0007-0492-6053} \and Gabriel~E.~Lima\inst{3}\customorcidAuthor{0009-0009-7599-8550} \and \\ David~Menotti\inst{3}\customorcidAuthor{0000-0003-2430-2030}\hspace{0.05mm} \and
    Rayson~Laroca\inst{1,3,*}\customorcidAuthor{0000-0003-1943-2711}
    }
\else
    \author{Anonymous Authors\inst{1,2,3,*}
}
\fi
\authorrunning{V. Orrú et al.}
\iffinal
    \institute{
    Pontifical Catholic University of Paraná, Curitiba, Brazil \and
    National High Court of Brazil, Brasília, Brazil \and
    Federal University of Paraná, Curitiba, Brazil \\[1.75ex]
    * \email{rayson@ppgia.pucpr.br}
    }
\else
    \institute{
    Anonymous Affiliation 1 \and
    Anonymous Affiliation 2 \and
    Anonymous Affiliation 3 \\[1.75ex]
    * \email{abc@def.ghi}
    }
\fi
\maketitle              %

\ifarxiv
    \copyrightnotice
\else
\fi

\newacronym{alpr}{ALPR}{Automatic License Plate Recognition}
\newacronym{cnn}{CNN}{Convolutional Neural Network}
\newacronym{fgvc}{FGVC}{Fine-Grained Vehicle Classification}
\newacronym{lwcd}{LWCD}{Linear
Warmup with Cosine Decay}
\newacronym{its}{ITS}{Intelligent Transportation Systems}
\newacronym{lp}{LP}{License Plate}
\newacronym{lpd}{LPD}{License Plate Detection}
\newacronym{lpr}{LPR}{License Plate Recognition}
\newacronym{senatran}{SENATRAN}{Brazilian National Traffic Secretariat}
\newacronym{sibgrapi}{SIBGRAPI}{Conference on Graphics, Patterns, and Images}
\newacronym{srr}{SRR}{Softmax Response Rejection}
\newacronym{spp}{SPP}{Spatial Pyramid Pooling}
\newacronym{ocr}{OCR}{Optical Character Recognition}
\newacronym{vcr}{VCR}{Vehicle Color Recognition}
\newacronym{vmmr}{VMMR}{Vehicle Make and Model Recognition}
\newacronym{vtr}{VTR}{Vehicle Type Recognition}
\newacronym{vit}{ViT}{Vision Transformer}
\newacronym{mm}{MM}{Multimodal}
\newacronym{mlp}{MLP}{Multi-Layer Perceptron}
\newacronym{wce}{WCE}{Weighted Cross-Entropy}
\newacronym{ce}{CE}{Cross-Entropy}
\newacronym{cd}{CD}{Cosine Decay}
\newacronym{pp}{pp}{percentage point}

\newcommand{\rodosolalpr}{RodoSol-ALPR\xspace}
\newcommand{\ufpralpr}{UFPR-ALPR\xspace}
\newcommand{\dataset}{UFPR-VeSV\xspace}

\begin{abstract}

Vehicle color recognition is an important cue for vehicle identification in surveillance systems, especially when license plates are illegible due to low resolution, occlusion, motion blur, or poor illumination. However, real-world vehicle color distributions are highly imbalanced, making overall accuracy insufficient to assess performance on rare but operationally relevant colors. This paper presents a comprehensive study of vehicle color recognition under severe class imbalance using UFPR-VeSV, a challenging real-world surveillance dataset. We investigate synthetic minority-class augmentation through two off-the-shelf generative strategies: text-conditioned image generation with RunDiffusion/Juggernaut-XL and image-conditioned color editing with Gemini 2.0 Flash. The curated synthetic data are combined with modern visual representations, loss reweighting, learning-rate scheduling, color-safe augmentation, foreground-aware preprocessing, and ensemble fusion. The best-performing approach achieves 94.6\% micro accuracy and 79.7\% macro accuracy, improving macro accuracy by 8.2 percentage points over recent literature. A manual error analysis further shows that many remaining failures are visually ambiguous even for human annotators, highlighting the practical limits of color-based vehicle identification in unconstrained surveillance imagery. The generated images and source code are publicly available at \url{https://github.com/viniciusorru/vcr-synthetic}.

\keywords{Vehicle color recognition  \and Synthetic data \and Class imbalance.}

\end{abstract}

\section{Introduction}
\label{sec:introduction}

\setcounter{footnote}{0}
\glsresetall

Vehicle identification in camera networks is a central task in intelligent transportation systems, supporting traffic monitoring, vehicle search, and forensic investigation~\cite{he2024vehicle,laroca2025advancing,yi2025advances}.
Although vehicle identification is often associated with \gls*{alpr}~\cite{ismail2025automatic}, license plates may be illegible due to occlusion, motion blur, poor illumination, sensor degradation, or low resolution~\cite{nascimento2024enhancing,laroca2026competition,wojcik2026lplcv2}.
Complementary visual attributes extracted from the vehicle body are therefore important in practical scenarios~\cite{ni2021vehicle,li2025weakly,lima2026toward}.
Among them, vehicle color is particularly useful because it is intuitive for human operators, covers a larger image region than the license plate, and can support searches even when the license plate cannot be recognized~\cite{chen2014vehicle,hu2023vehicle,lima2024toward,lima2026toward}.

Early \gls*{vcr} methods relied mainly on handcrafted features, color-space analysis, and explicit vehicle-region detection~\cite{baek2007vehicle,chen2014vehicle,hsieh2015vehicle}. 
More recently, deep learning methods have achieved strong results on controlled benchmarks~\cite{hu2015vehicle,fu2018mcff,wang2021transformer}. 
However, real-world surveillance imagery remains challenging due to heterogeneous viewpoints, adverse weather, changing illumination, glare, motion blur, sensor noise, nighttime acquisition, and partial occlusions. 
These factors directly affect the visual evidence available for color recognition, motivating recent datasets and methods designed to evaluate \gls*{vcr} under more realistic conditions~\cite{hu2022rain,hu2023vehicle,lima2024toward,lima2026toward}.

A further difficulty is the severe class imbalance naturally found in vehicle color distributions. 
In real traffic, neutral colors such as white, black, silver, and gray are far more frequent than colors such as yellow, orange, green, beige, and brown~\cite{iseecars2024cars,ministeri2025shades}. 
As a consequence, a model may achieve high overall accuracy while still performing poorly on rare colors, since micro-level performance is dominated by frequent classes. 
This limitation is particularly relevant in surveillance and forensic applications, where uncommon colors can be highly informative for reducing the search space. 
Therefore, \gls*{vcr} should not be assessed only by overall accuracy, but also by macro-level metrics that give equal importance to each class and better expose failures on underrepresented~colors.

Prior \gls*{vcr} studies have addressed class imbalance mainly through re-weighted losses, balanced sampling, and conventional data augmentation~\cite{hu2023vehicle,lima2024toward}. 
Although these strategies are useful, they remain constrained by the diversity of the available real images. 
For minority classes, the training samples may cover only a limited range of vehicle types, viewpoints, illumination conditions, backgrounds, and acquisition artifacts. 
This motivates the use of generative models as a complementary way to increase visual support for rare colors. 
Recent studies have shown that diffusion-based synthetic images can improve image classification when generated and filtered appropriately~\cite{azizi2023synthetic,trabucco2024effective}. 
Nevertheless, in practical \gls*{vcr}, synthetic data should be evaluated as part of a broader recognition pipeline, together with imbalance-aware training, robust representations, suitable augmentation, and careful error analysis.

In this work, we study \gls*{vcr} on \dataset~\cite{lima2026toward}, a recently introduced and highly diverse real-world vehicle surveillance dataset.
Rather than presenting synthetic data generation as the main contribution by itself, we provide a comprehensive study of strategies for improving minority-class recognition in \gls*{vcr}.
We investigate two complementary off-the-shelf generation strategies: text-conditioned generation with RunDiffusion/Juggernaut-XL~\cite{rundiffusion_juggernaut_xl} and image-conditioned editing with Gemini 2.0 Flash \cite{google2025gemini20flashimage}.
The generated images are manually filtered through an inter-rater quality assessment protocol and combined with several recognition strategies, including DINOv3 features~\cite{simeoni2025dinov3}, loss reweighting, color-safe augmentation, background segmentation, and ensemble fusion.

Our contributions are threefold.
First, we provide an extensive evaluation of modern recognition strategies for \gls*{vcr} under severe class imbalance, focusing on macro-level performance rather than only on overall accuracy.
Second, we build and evaluate two curated synthetic image sets designed to increase the visual support for minority color classes.
Third, we conduct a detailed error analysis of the final predictions, distinguishing cases that appear correctable by the model from those that remain ambiguous even for human annotators.
To the best of our knowledge, this is the first \gls*{vcr} study to analyze the remaining errors at this level of detail, providing a clearer view of both the practical value and the limitations of vehicle color as a surveillance cue.

The remainder of this paper is organized as follows.
\cref{sec:dataset} presents \dataset and the proposed synthetic minority-class augmentation process.
\cref{sec:setup} describes the experimental setup.
\cref{sec:results} reports the results and error analysis.
Finally, \cref{sec:conclusions} concludes the paper and outlines future research directions.

\section{UFPR-VeSV and Synthetic Augmentation}
\label{sec:dataset}

\subsection{Reference Dataset and Class Imbalance}

We use \dataset~\cite{lima2026toward}, a real-world surveillance dataset for unified fine-grained vehicle classification and \gls*{alpr}.
It contains operational-camera images acquired under heterogeneous viewpoints, distances, illumination, weather, motion blur, nighttime infrared capture, and partial occlusions, making it suitable for studying \gls*{vcr} beyond near-ideal imagery.

\begin{figure}[!htb]
    \centering
    \includegraphics[width=0.95\linewidth]{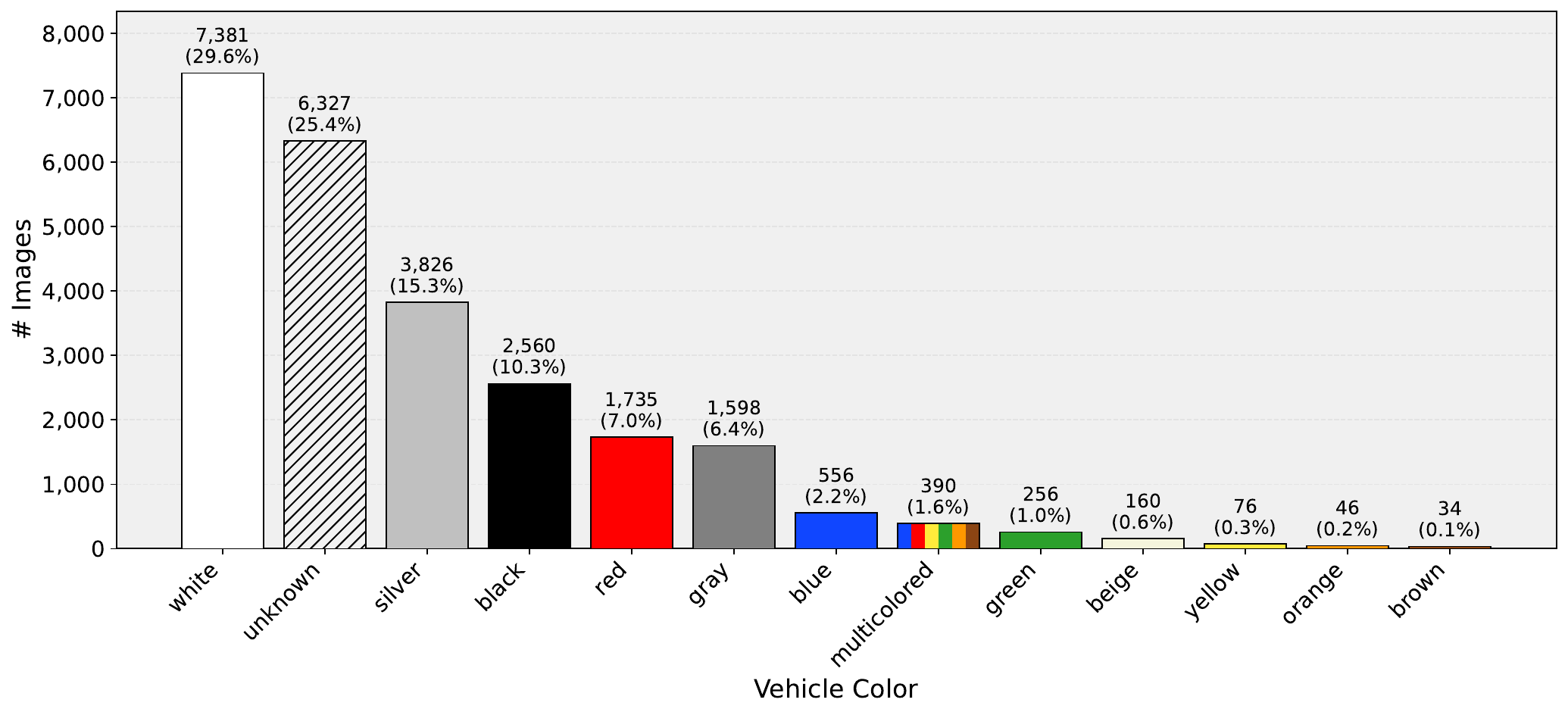}

    \vspace{-4mm}
    
    \caption{Distribution of vehicle colors in \dataset~\cite{lima2026toward}. The strong imbalance motivates macro-level evaluation and synthetic data focused on minority classes.}
    \label{fig:distribution_ufpr_vesv}
\end{figure}

The dataset comprises 24,945 images of 16,297 unique vehicles collected from the Military Police of Paran\'{a}, Brazil.
Images are annotated with 13 color classes and other attributes not explored here.
As shown in \cref{fig:distribution_ufpr_vesv}, the distribution is highly long-tailed: white contains 7,381 images, whereas brown contains only~34.
The ``unknown'' class mainly corresponds to infrared nighttime images and motorcycle samples for which color cannot be reliably determined; it is kept in the recognition task but is not used as a target for synthetic augmentation.

Lima et al.~\cite{lima2026toward} reported strong overall results on \dataset, but the best color-recognition configuration reached only 71.5\% macro accuracy.
This gap motivates our analysis of whether synthetic images, imbalance-aware training, and modern visual representations can improve minority-class prediction.

\subsection{Synthetic Image Generation Strategy}

Our synthetic data are intended to increase the number and diversity of training samples for underrepresented colors, not to replace real surveillance images.
We consider two complementary strategies: text-to-image generation, which can create diverse vehicles, scenes, viewpoints, and lighting conditions; and image-conditioned editing, which changes vehicle color while preserving much of the original structure and scene context.
RunDiffusion/Juggernaut-XL follows the first strategy, using a photorealistic model derived from Stable Diffusion XL~\cite{podell2024sdxl,rundiffusion_juggernaut_xl}, while Gemini 2.0 Flash image generation supports multimodal image editing from image and text inputs~\cite{google2025gemini20flashimage}.
Both models are used off the shelf, without fine-tuning.

\subsection{Text-to-Image Generation with RunDiffusion/Juggernaut-XL}

The RunDiffusion subset was generated with the Juggernaut-XL checkpoint~\cite{rundiffusion_juggernaut_xl}.
For each image, the script sampled a target color by favoring underrepresented classes in the original training distribution and reducing the probability of classes already frequent among the generated candidates.
It also selected a common Brazilian vehicle model with inverse-frequency weighting.
Prompts combined vehicle make/model, target color, location, environmental condition, viewpoint, and distance, including Brazilian roads and urban scenes under sunny, cloudy, rainy, nighttime, sunset, and foggy conditions.
For the multicolored class, the prompt explicitly required vehicles with multiple~colors.

\begin{figure}[!htb]
    \centering

    \resizebox{0.925\linewidth}{!}{
        \includegraphics[height=13ex]{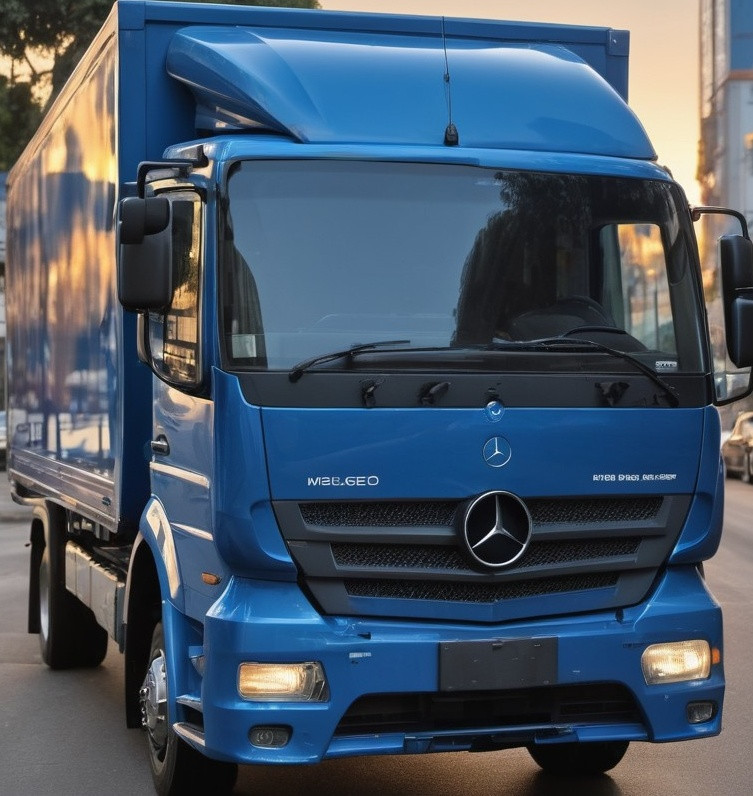}
        \includegraphics[height=13ex]{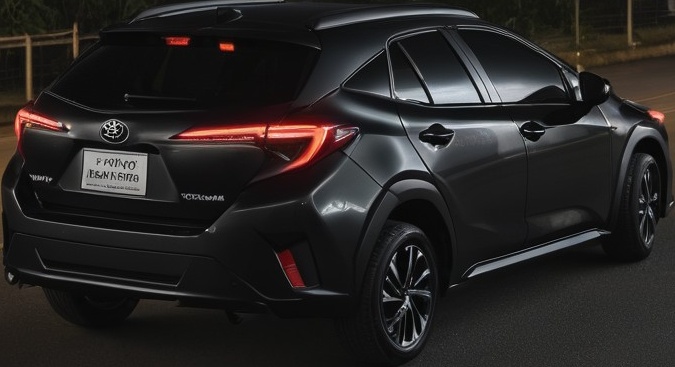}
        \includegraphics[height=13ex]{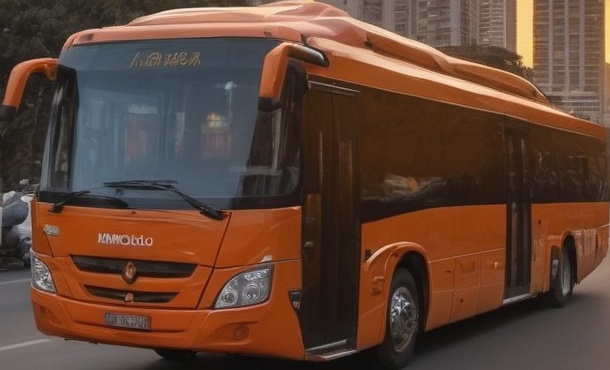}
        \includegraphics[height=13ex]{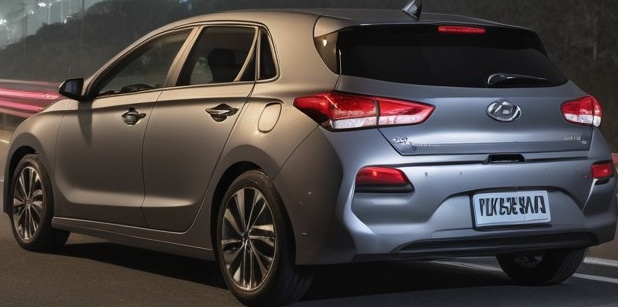}
    }

    \vspace{0.5mm}

    \resizebox{0.925\linewidth}{!}{
        \includegraphics[height=11.5ex]{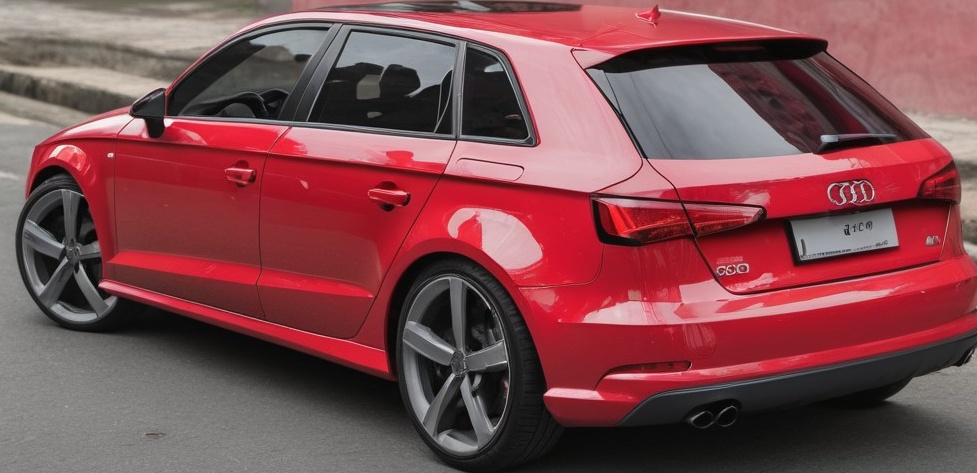}
        \includegraphics[height=11.5ex]{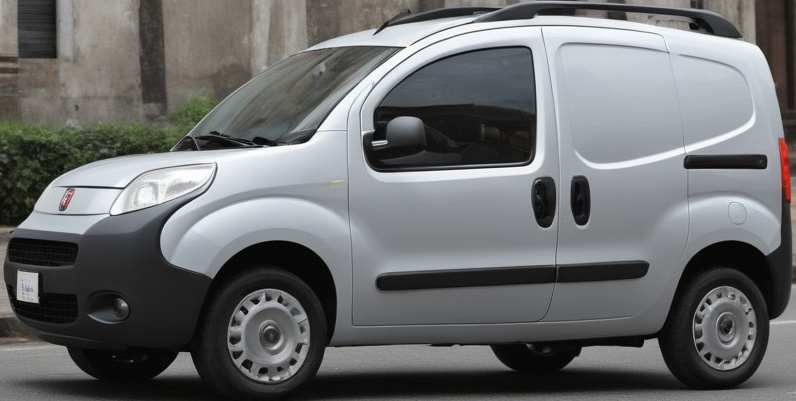}
        \includegraphics[height=11.5ex]{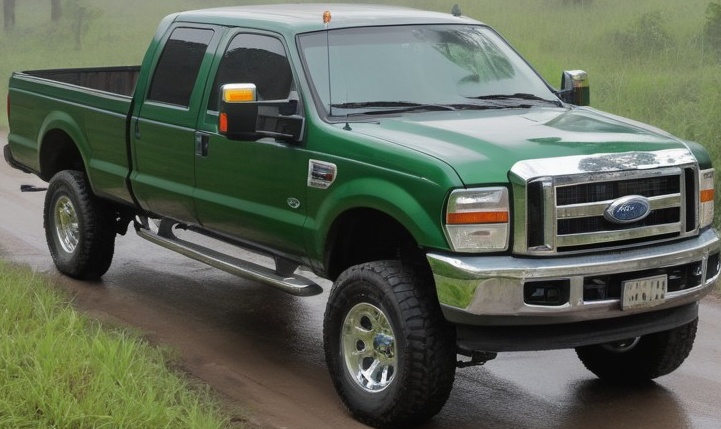}
        \includegraphics[height=11.5ex]{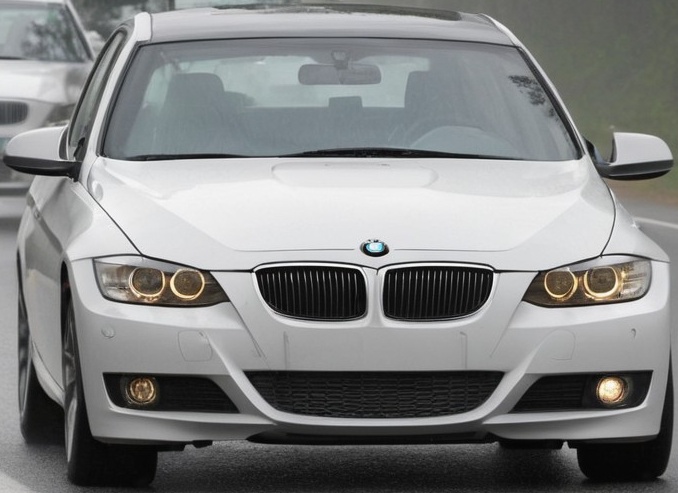}
    }

    \vspace{0.5mm}
    
    \resizebox{0.925\linewidth}{!}{
        \includegraphics[height=12ex]{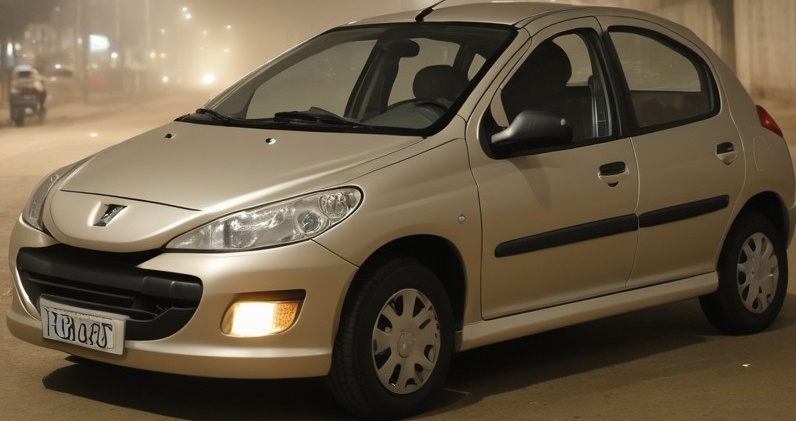}
        \includegraphics[height=12ex]{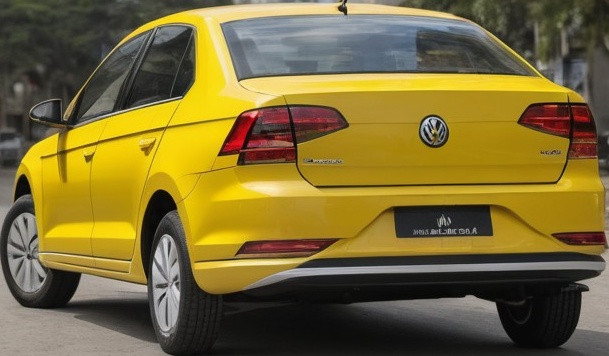}
        \includegraphics[height=12ex]{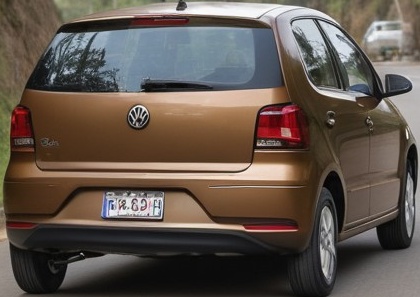}
        \includegraphics[height=12ex]{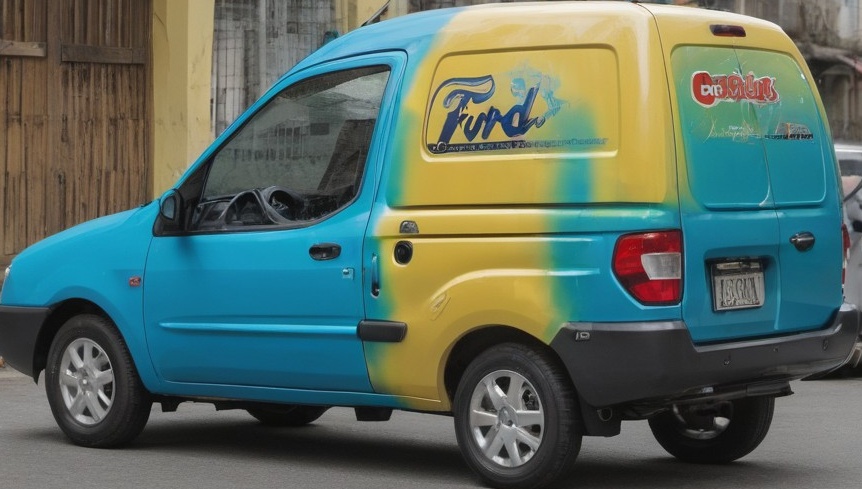}
    }

    \vspace{-1.75mm}
    
    \caption{Representative images generated with RunDiffusion/Juggernaut-XL~\cite{rundiffusion_juggernaut_xl}. Samples are arranged from left to right and top to bottom as blue, black, orange, silver, red, gray, green, white, beige, yellow, brown, and multicolored.}
    \label{fig:samples-run_diffusion}
\end{figure}

After generation, YOLOv11~(small)~\cite{yolov11} was used to detect the main vehicle.
Only the largest vehicle-related bounding box was stored with the sample metadata.
\cref{fig:samples-run_diffusion} shows representative examples, illustrating the intended diversity of vehicles, viewpoints, and scenes for minority colors.

In total, 19,144 RunDiffusion images were generated.
After quality control (see \cref{sec:dataset:quality_control}) and duplicate removal, 12,951 images were retained.

\subsection{Image-Conditioned Color Editing with Gemini 2.0 Flash}

\begin{figure}[!htb]
    \centering
    \captionsetup[subfigure]{captionskip=1pt,labelformat=empty,font=scriptsize}

    \resizebox{0.99\linewidth}{!}{
    \subfloat[Original $\rightarrow$ Red]{
        \includegraphics[height=11.25ex]{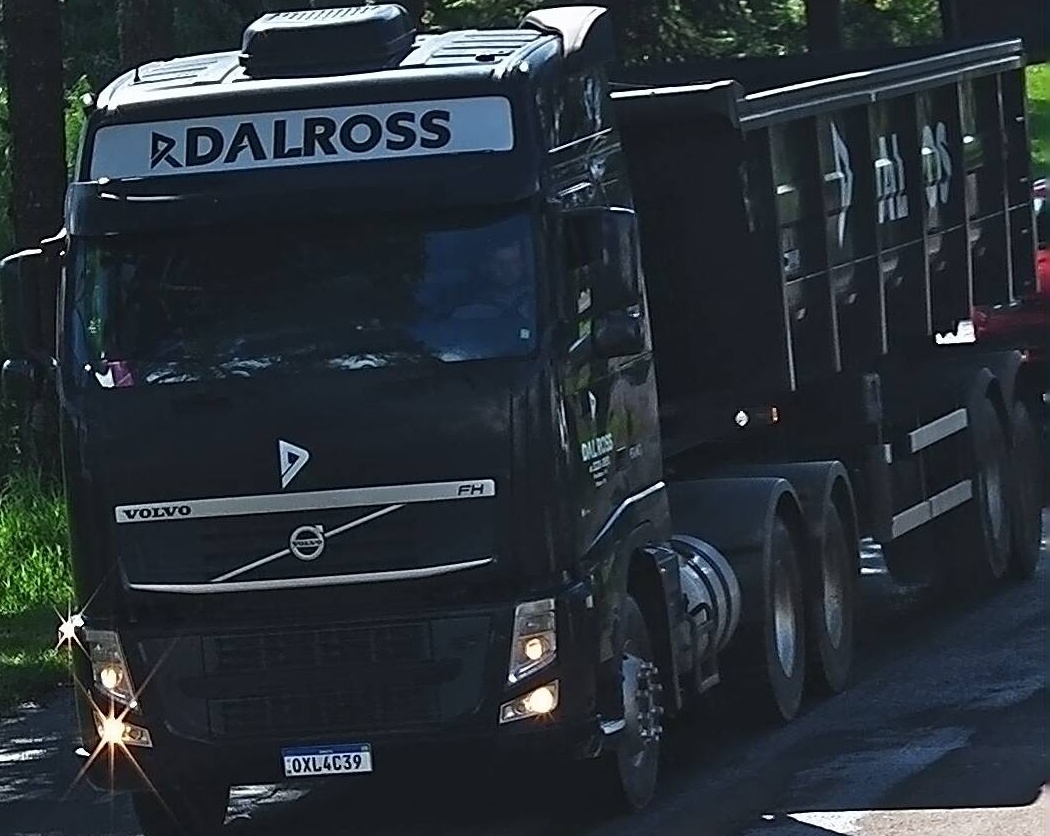}
        \includegraphics[height=11.25ex]{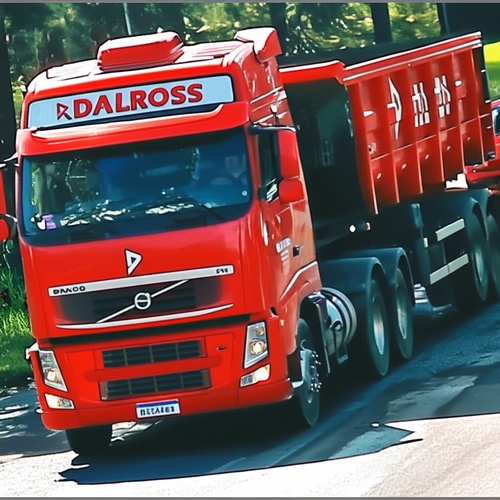}
    } \quad
    \subfloat[Original $\rightarrow$ Orange]{
        \includegraphics[height=11.25ex]{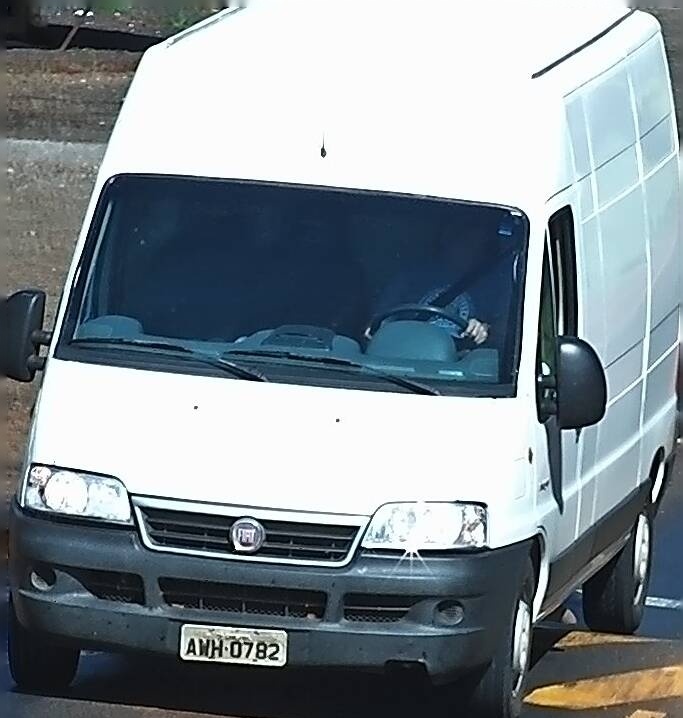}
        \includegraphics[height=11.25ex]{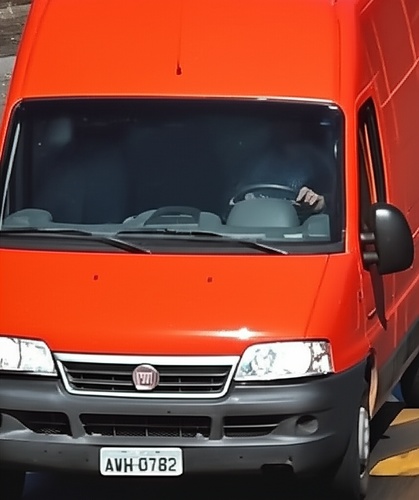}
    } \quad
    \subfloat[Original $\rightarrow$ Brown]{
        \includegraphics[height=11.25ex]{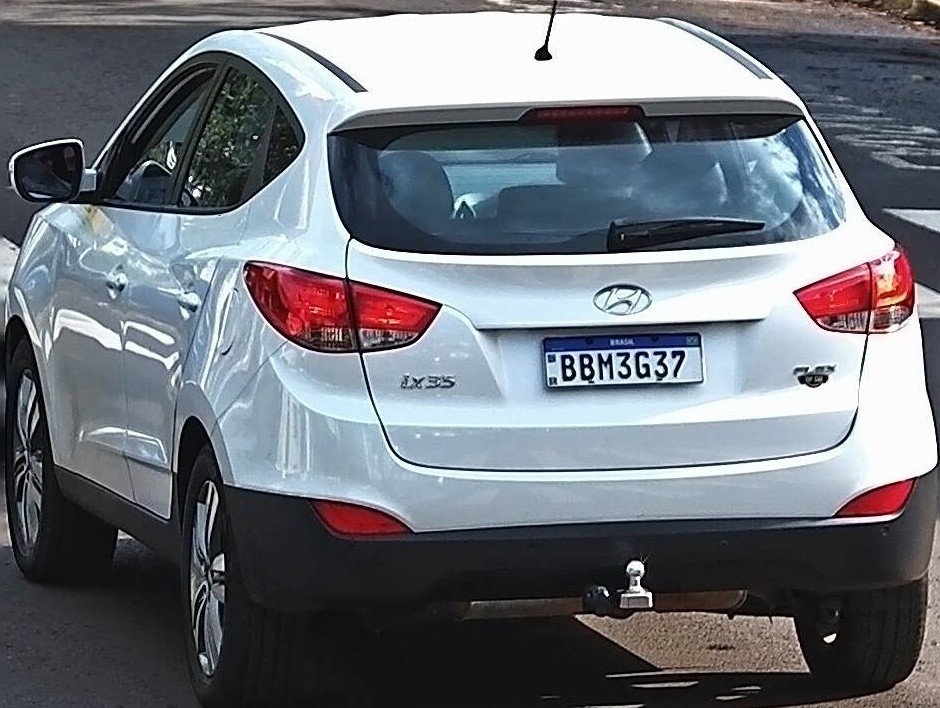}
        \includegraphics[height=11.25ex]{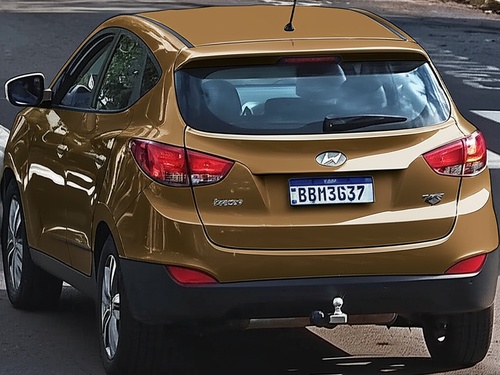}
    }
    }

    \vspace{2.25mm}

    \resizebox{0.99\linewidth}{!}{
    \subfloat[Original $\rightarrow$ Yellow]{
        \includegraphics[height=13.25ex]{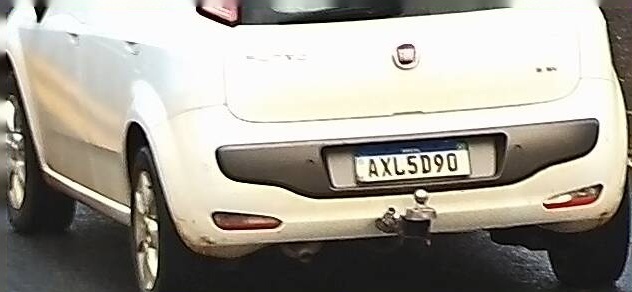}
        \includegraphics[height=13.25ex]{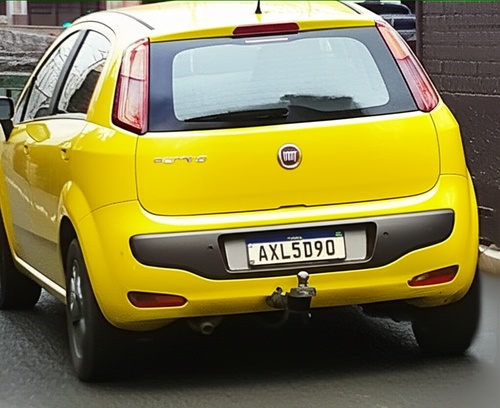}
    } \quad
    \subfloat[Original $\rightarrow$ Red]{
        \includegraphics[height=13.25ex]{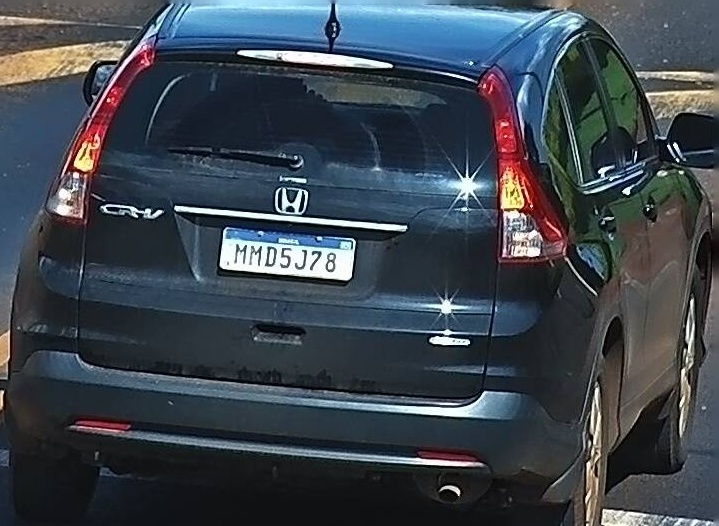}
        \includegraphics[height=13.25ex]{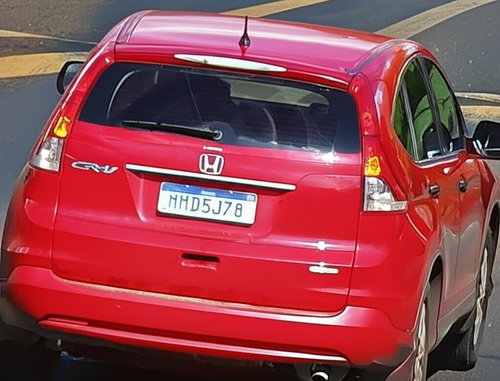}
    }
    }

    \vspace{-1.75mm}

    \caption{Representative images generated with Gemini. Each pair shows the original image from \dataset~\cite{lima2026toward} on the left and the generated color variant on the right.}
    \label{fig:gemini_samples}
\end{figure}

Gemini receives an existing image from \dataset along with a textual instruction, generating a modified version that changes vehicle body color while preserving the camera geometry, vehicle pose, background, and image quality.
For each selected source image and target color, the prompt asked Gemini to change the entire vehicle body to the target color, avoid vibrant tones, keep realistic vehicle colors, and preserve a visible front or rear view.
Due to API rate limits, 14,040 Gemini images were generated, of which 6,028 were accepted after quality control.
\cref{fig:gemini_samples} illustrates the image-conditioned nature of this strategy.

\subsection{Manual Quality Control and Inter-Rater Agreement}
\label{sec:dataset:quality_control}

All synthetic images were evaluated by two independent annotators before training.
The criteria were visual realism, correctness of the target color, and consistency with the vehicle surveillance domain.
For RunDiffusion, annotators accepted realistic, non-deformed vehicles with the intended color; for Gemini, edited samples also had to preserve plausible vehicle appearance without visible editing artifacts.
Only images accepted by both annotators were retained.

\cref{tab:interrater} reports the observed agreement~($P_o$), i.e., the proportion of cases where both annotators agreed on the same label, alongside Cohen's kappa~({\large$\kappa$})~\cite{cohen1960coefficient}, which adjusts for the agreement expected by chance alone. The resulting {\large$\kappa$} values fall in the moderate range of the Landis-Koch scale~\cite{landis1977measurement}, which is consistent with the inherent ambiguity of borderline color categories such as gray~$\leftrightarrow$~silver, beige~$\leftrightarrow$~brown, and colors captured under low-light~conditions.

\begin{table}[!htb]
\centering
\setlength{\tabcolsep}{6pt}
\caption{Inter-rater agreement for synthetic image evaluation.}
\label{tab:interrater}

\vspace{-2.25mm}

\resizebox{0.775\linewidth}{!}{
\begin{tabular}{@{}lcccc@{}}
\toprule
Source & $N$ & $P_o$ & {\large$\kappa$} & Accepted \\
\midrule
RunDiffusion/Juggernaut-XL     & 19,144 & 0.80 & 0.43 & 12,951 (67.7\%) \\
Gemini 2.0 Flash                 & 14,040 & 0.71 & 0.41 & \phantom{0}6,028 (42.9\%) \\
\bottomrule
\end{tabular}
}

\end{table}

\subsection{Final Augmented Dataset Composition}
\label{sec:augmented_composition}

\cref{tab:class_distribution} reports the number of original and accepted synthetic images per color class.
The synthetic sets mainly target minority colors, especially blue, green, yellow, orange, and brown.
The final distribution also reflects the quality-control stage: some minority classes, particularly multicolored and beige, received fewer accepted samples than expected.
Multicolored vehicles were often generated with a single dominant color or patterns that were difficult to label consistently, while beige frequently drifted toward visually similar classes such as white, silver, gray, or brown.
These borderline cases likely reflect low visual separability, biases in the generative models' training data, and increased annotator disagreement.
Majority classes are augmented less aggressively, and the ``unknown'' class is not augmented because its label indicates unreliable color~evidence.

\begin{table}[!htb]
\centering
\setlength{\tabcolsep}{6pt}
\caption{Color class distribution in the original \dataset dataset and in the accepted synthetic subsets. The ``unknown'' class corresponds mainly to infrared nighttime images and motorcycle samples for which color cannot be reliably determined.}
\label{tab:class_distribution}

\vspace{-2.25mm}

\resizebox{0.85\linewidth}{!}{
\begin{tabular}{@{}lcccr@{}}
\toprule
Color & \phantom{i}\dataset & \phantom{0}RunDiffusion & \phantom{0}Gemini & \multicolumn{1}{c@{}}{Total} \\
\midrule
White        & \phantom{0}7,381 &   \phantom{00,}173 &   \phantom{0,}581 & \phantom{0}8,135\phantom{00,} (+10.2\%) \\
Unknown      & \phantom{0}6,327 &  \phantom{00}$-$  &  \phantom{0}$-$  & \phantom{0}6,327\phantom{00,} (+\phantom{0}0.0\%) \\
Silver       & \phantom{0}3,826 &   \phantom{00,}159 &   \phantom{0,}267 & \phantom{0}4,252\phantom{00,} (+11.1\%) \\
Black        & \phantom{0}2,560 &   \phantom{00,}170 &   \phantom{0,}488 & \phantom{0}3,218\phantom{00,} (+25.7\%) \\
Red          & \phantom{0}1,735 &   \phantom{00,}428 &   \phantom{0,}886 & \phantom{0}3,049\phantom{00,} (+75.7\%) \\
Gray         & \phantom{0}1,598 &   \phantom{00,}640 &   \phantom{0,}163 & \phantom{0}2,401\phantom{00,} (+50.3\%) \\
Blue         &   \phantom{00,}556 & \phantom{0}2,001 &   \phantom{0,}879 & \phantom{0}3,436\phantom{0,} (+518.0\%) \\
Multicolored &   \phantom{00,}390 &   \phantom{00,}640 &   \phantom{0,}438 & \phantom{0}1,468\phantom{0,} (+276.4\%) \\
Green        &   \phantom{00,}256 & \phantom{0}2,135 &   \phantom{0,}718 & \phantom{0}3,109 (+1{,}114.5\%) \\
Beige        &   \phantom{00,}160 &   \phantom{00,}365 &   \phantom{0,}123 &   \phantom{00,}648\phantom{0,} (+305.0\%) \\
Yellow       &    \phantom{000,}76 & \phantom{0}1,946 &   \phantom{0,}565 & \phantom{0}2,587 (+3{,}303.9\%) \\
Orange       &    \phantom{000,}46 & \phantom{0}2,129 &   \phantom{0,}683 & \phantom{0}2,858 (+6{,}113.0\%) \\
Brown        &    \phantom{000,}34 & \phantom{0}2,165 &   \phantom{0,}237 & \phantom{0}2,436 (+7{,}064.7\%) \\
\midrule
\textbf{Total} & \textbf{24,945} & \textbf{12,951} & \textbf{6,028} & \textbf{43,924\phantom{,} (+76.1\%)} \\
\bottomrule
\end{tabular}
}
\end{table}

\section{Experimental Setup}
\label{sec:setup}

\subsection{Evaluation Protocol}
\label{sec:metrics}

We follow the stratified 3:1:1 train--validation--test protocol of Lima et al.~\cite{lima2026toward}, using ten independent splits generated with the script released by the authors.
This preserves the evaluation setup and enables direct comparison with the original \dataset benchmark.
All results are reported exclusively on real test images; synthetic images are reserved for training only. RunDiffusion/Juggernaut-XL images are appended directly to each training split, while Gemini-generated images are included only when their source image belongs to the corresponding training partition, thereby preventing source-level data leakage~\cite{laroca2022first,laroca2023do}.

The experiments consider four training-set configurations: the original \dataset training data, \dataset + RunDiffusion, \dataset + Gemini, and \dataset + Gemini + RunDiffusion.
Following prior work on \gls*{vcr}~\cite{lima2024toward,lima2026toward}, we report micro accuracy~(Mi-Acc), macro accuracy~(Ma-Acc), and macro F1-score~(F1), averaged over the ten splits.
Among these, we focus primarily on macro accuracy, as it directly reflects per-class performance and is straightforward to interpret in the context of long-tailed~distributions.

\subsection{Recognition Models and Experimental Design}

To avoid tailoring conclusions to a single architecture, we evaluate four supervised baselines used in recent \gls*{vcr} studies~\cite{lima2024toward,lima2026toward}: EfficientNet-V2~\cite{tan2021efficientnetv2}, ResNet-50~\cite{he2016residual}, Swin Transformer Tiny~\cite{liu2022swin}, and ViT-B/16~\cite{dosovitskiy2021vit}, all initialized with ImageNet-pretrained weights.

We also evaluate DINOv3~\cite{simeoni2025dinov3} as a frozen self-supervised feature extractor, motivated by its transferable representations under limited supervision.
Following prior evidence that frozen DINOv2 features can be effective when the target domain remains close to natural imagery~\cite{huang2024comparative}, we test the Small~(ViT-S/16, 21M parameters), Base~(ViT-B/16, 86M), and Large~(ViT-L/16, 300M) variants.
For each model, the frozen \texttt{[CLS]} token feeds a two-layer \gls*{mlp} head with a 512-dimensional projection, batch normalization, ReLU, dropout~($p=0.4$), and a final~classifier.

\subsection{Optimization and Class-Imbalance Handling}

All models are optimized with Adam~\cite{kingma2015adam} and trained for up to 100 epochs with early stopping.
We compare standard \gls*{ce} with \gls*{wce}, a common strategy for imbalanced recognition~\cite{huang2016learning,zhang2023deep}.
We also evaluate three learning-rate schedules: \gls*{cd}~\cite{loshchilov2017sgdrstochasticgradientdescent}, \gls*{lwcd}, and ReduceLROnPlateau.
For \gls*{lwcd}, the learning rate increases from $\eta_{\min}=10^{\text{-}6}$ to $\eta_{\max}=10^{\text{-}4}$ during the first five epochs and then follows \gls*{cd}~\cite{goyal2017accurate}.
DINOv3 experiments use batch size 32, learning rate $10^{\text{-}3}$, and weight decay $10^{\text{-}5}$ due to GPU memory constraints.
All experiments were conducted on an NVIDIA RTX~5090 GPU with 32~GB of memory.

\subsection{Color-Safe Augmentation and Foreground Preprocessing}
Data augmentation must be especially conservative when color is the target attribute. We apply an Albumentations~\cite{buslaev2020albumentations} pipeline with probability 0.8, randomly sampling three transformations from affine transformations ($\pm15^{\circ}$ rotation and scale $0.9$--$1.1$), horizontal flip, Gaussian noise, blur, brightness/contrast variation ($\pm20\%$), and conservative HSV perturbations. To avoid corrupting the color signal, hue, saturation, and value shifts are limited to $\pm5$ levels~($\approx2.8\%$ of the OpenCV range), $\pm20$ levels~($\approx7.8\%$), and $\pm10$ levels~($\approx3.9\%$ of the 8-bit range), respectively.
We also carefully inspected augmented samples throughout this process to ensure that the perceived color was~preserved.

\begin{figure}[!htb]
    \vspace{-2mm}
    
    \centering
    \resizebox{0.99\linewidth}{!}{
        \includegraphics[height=15ex]{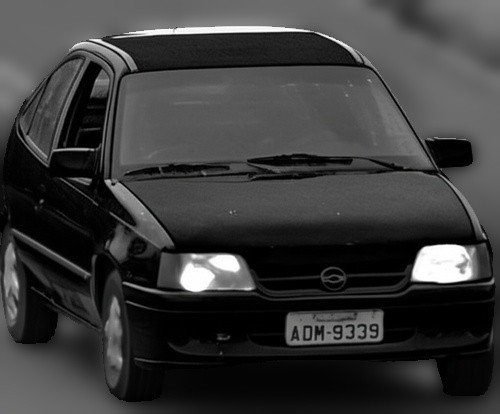}
        \includegraphics[height=15ex]{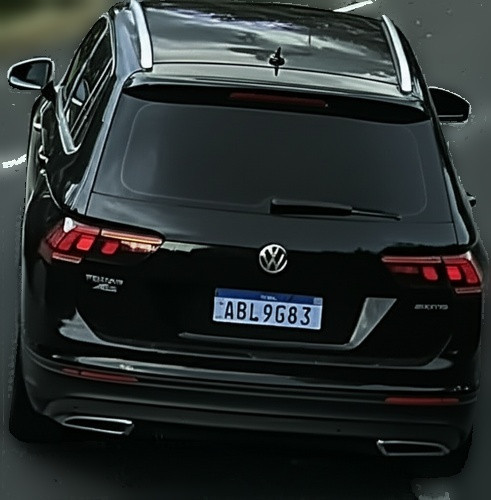}   
        \includegraphics[height=15ex]{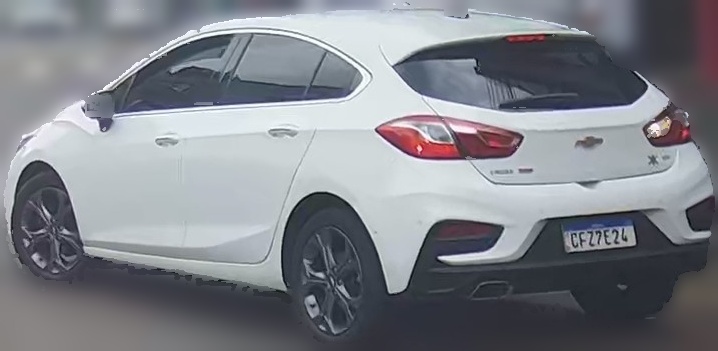}
        \includegraphics[height=15ex]{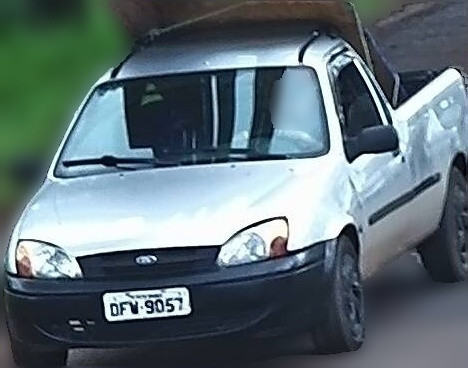}
    }

    \vspace{-2mm}
    \caption{Examples of foreground-aware preprocessing with SAM~2~\cite{ravi2024sam2segmentimages}.
    }
    \label{fig:segmented_examples}

    \vspace{-2mm}
\end{figure}

We also evaluate a foreground-aware preprocessing strategy based on SAM~2~\cite{ravi2024sam2segmentimages}.
In this setting, the vehicle region is preserved while the background is blurred, as illustrated in \cref{fig:segmented_examples}.
This experiment assesses whether emphasizing vehicle paint helps reduce the model's dependence on contextual cues such as the scene background, shadows, and road~surface.

\subsection{Model Fusion}

Finally, we evaluate whether complementary models improve robustness on minority classes.
For each split, the top three candidates are selected based on validation macro accuracy and then combined by hard voting.
When the three models disagree, the prediction with the highest class confidence is used only as a tiebreaker.
Importantly, we also evaluated a confidence-based fusion rule, which directly selects the prediction with the highest class confidence, but it performed slightly below hard voting and the difference was not statistically significant.
Therefore, we focus on hard voting in the remaining~analysis.

\section{Results and Discussion}
\label{sec:results}

\subsection{Overall Performance}

\cref{tab:baseline} compares the performance of the previous \dataset benchmark~\cite{lima2026toward} against our baselines and the final proposed approach.
Among the single models trained exclusively on the original data, DINOv3-Large achieves the most favorable balance between overall and class-specific performance, reaching 94.0\% micro-accuracy and 72.5\% macro-accuracy.
While this represents a 1.0~\gls*{pp} improvement in macro-accuracy over the results reported by Lima et al.~\cite{lima2026toward}, this specific gain is not statistically significant~($p > 0.05$).

\begin{table}[!htb]
\centering
\setlength{\tabcolsep}{6pt}
\caption{Baseline results on the original \dataset dataset and performance of the proposed approach. The baselines use \gls*{ce} loss, \gls*{cd}, and no data augmentation. The proposed approach combines Gemini-augmented training data, \gls*{wce}, \gls*{lwcd}, color-safe augmentation, foreground-aware preprocessing, and hard-voting ensemble fusion.}
\label{tab:baseline}

\vspace{-2.25mm}

\resizebox{0.735\linewidth}{!}{
\begin{tabular}{@{}lccc@{}}
\toprule
Model & Mi-Acc (\%) & Ma-Acc (\%) & F1 (\%) \\
\midrule
EfficientNet-V2 (from~\cite{lima2026toward}) & 93.5 $\pm$ 0.6 & 71.5 $\pm$ 2.3 & 73.8 $\pm$ 2.3 \\
\midrule
ResNet-50             & 91.2 $\pm$ 0.7 & 64.1 $\pm$ 2.3 & 66.4 $\pm$ 2.2 \\
EfficientNet-V2       & 90.9 $\pm$ 1.0 & 66.7 $\pm$ 2.5 & 68.2 $\pm$ 1.9 \\
ViT-B/16              & 92.8 $\pm$ 0.6 & 69.9 $\pm$ 3.4 & 72.5 $\pm$ 2.8 \\
Swin-T                & 93.2 $\pm$ 0.6 & 72.5 $\pm$ 2.9 & 74.0 $\pm$ 2.8 \\
\midrule
DINOv3-Small          & 92.5 $\pm$ 0.6 & 69.1 $\pm$ 3.1 & 71.2 $\pm$ 2.2 \\
DINOv3-Base           & 93.0 $\pm$ 0.4 & 69.6 $\pm$ 3.4 & 72.3 $\pm$ 3.1 \\
DINOv3-Large          & 94.0 $\pm$ 0.5 & 72.5 $\pm$ 3.2 & 74.4 $\pm$ 2.5 \\
\midrule
\textbf{Proposed approach} & \textbf{94.6 $\pm$ 0.4} & \textbf{79.7 $\pm$ 4.0} & \textbf{78.8 $\pm$ 3.1} \\
\bottomrule
\end{tabular}
}
\end{table}

In contrast, our full proposed pipeline reaches 79.7\% macro-accuracy—a significant improvement of 7.2~\gls*{pp} over our DINOv3-Large baseline and 8.2~\gls*{pp} over the previous benchmark ($p < 0.05$).
As detailed in \cref{sec:results:ablation}, these gains are cumulative, resulting from the integration of Gemini-augmented data, loss reweighting, \gls*{lwcd}, color-safe augmentation, foreground-aware preprocessing, and hard-voting ensemble fusion. Notably, while the increase in micro-accuracy is more modest (from 93.5\% to 94.6\%), it remains statistically significant ($p < 0.05$).
This suggests that our methodology primarily enhances performance on minority classes and visually challenging colors, improvements that are more effectively captured by macro-level metrics.

\subsection{Effect of Synthetic Data and Loss Reweighting}

\cref{tab:ce_wce_datasets} isolates the effect of training-set composition under \gls*{ce} and \gls*{wce}, using \acrfull*{cd} without data augmentation.
To keep the analysis concise, we report the best result obtained for each configuration; the same trend holds across model architectures.

\begin{table}[!htb]
\centering
\setlength{\tabcolsep}{5pt}
\caption{Test results for each training-set composition under fixed \gls*{ce} and \gls*{wce} settings, using \gls*{cd}, no data augmentation, and the 10-fold average.}
\label{tab:ce_wce_datasets}

\vspace{-2mm}

\resizebox{0.85\linewidth}{!}{
\begin{tabular}{@{}lccc@{}}
\toprule
Training configuration & Mi-Acc (\%) & Ma-Acc (\%) & F1 (\%) \\
\midrule
\multicolumn{4}{@{}l}{\textit{Cross-Entropy (CE)}} \\
\dataset & 94.0 $\pm$ 0.5 & 72.5 $\pm$ 3.2 & 74.4 $\pm$ 2.5 \\
\dataset + RunDiffusion & 93.5 $\pm$ 0.5 & 72.5 $\pm$ 4.6 & 74.6 $\pm$ 4.4 \\
\textbf{\dataset + Gemini} & \textbf{93.8 $\pm$ 0.3} & \textbf{74.9 $\pm$ 4.7} & \textbf{76.4 $\pm$ 3.5} \\
\dataset + Gemini + RunDiffusion & 93.1 $\pm$ 0.7 & 72.8 $\pm$ 3.8 & 74.1 $\pm$ 2.8 \\
\midrule
\multicolumn{4}{@{}l}{\textit{Weighted Cross-Entropy (WCE)}} \\
\dataset & 36.4 $\pm$ 0.3 & 59.3 $\pm$ 3.0 & 55.8 $\pm$ 3.3 \\
\dataset + RunDiffusion & 91.5 $\pm$ 0.4 & 71.5 $\pm$ 5.6 & 72.7 $\pm$ 5.4 \\
\textbf{\dataset + Gemini} & \textbf{92.7 $\pm$ 0.9} & \textbf{75.7 $\pm$ 3.0} & \textbf{75.0 $\pm$ 2.7} \\
\dataset + Gemini + RunDiffusion & 92.6 $\pm$ 0.1 & 71.9 $\pm$ 4.4 & 73.5 $\pm$ 3.5 \\
\bottomrule
\end{tabular}
}
\end{table}

Under \gls*{ce}, the differences are modest: the original dataset remains competitive, while Gemini achieves the highest macro accuracy, at 74.9\%.
This suggests that synthetic data alone brings limited gains when the objective is dominated by frequent classes.
The effect is clearer under \gls*{wce}.
On the original long-tailed data, reweighting is unstable and reduces macro accuracy to 59.3\%.
With synthetic images, \gls*{wce} becomes effective again, reaching 71.5\% with RunDiffusion, 75.7\% with Gemini, and 71.9\% with the combined set.
Gemini is strongest despite adding fewer images than RunDiffusion, suggesting that domain alignment is more important than volume alone.
The lower result of the combined set also indicates that adding more synthetic samples is not automatically beneficial when part of the data shifts the training distribution.

\subsection{Cumulative Ablation}
\label{sec:results:ablation}

\cref{tab:ablation} summarizes the best-performing path from the original DINOv3-Large baseline to the final ensemble.
The rows are a practical progression rather than isolated factorial effects, since the best model may change after each modification.
Still, each added component improves macro~accuracy.

\begin{table}[!htb]
\centering
\setlength{\tabcolsep}{4pt}
\caption{Best-performing path from the original baseline to the final ensemble. The last column reports the macro-accuracy gain relative to the previous row.}
\label{tab:ablation}

\vspace{-2.25mm}

\resizebox{0.99\linewidth}{!}{%
\begin{tabular}{@{}lllcccc@{}}
\toprule
Approach & Configuration & Best model & Mi-Acc (\%) & Ma-Acc (\%) & F1 (\%) & $\Delta$Ma \\
\midrule
Lima et al.~\cite{lima2026toward} & Original & EfficientNet-V2 & 93.5 $\pm$ 0.6 & 71.5 $\pm$ 2.3 & 73.8 $\pm$ 2.3 & \phantom{i}$-$ \\[0.3ex] \cdashline{1-7} \\[-1.75ex]
Our baseline & Original + \gls*{cd} & DINOv3-Large & 94.0 $\pm$ 0.5 & 72.5 $\pm$ 3.2 & 74.4 $\pm$ 2.5 & $-$ \\
+ Gemini data & Gemini + \gls*{cd} & DINOv3-Large & 93.8 $\pm$ 0.3 & 74.9 $\pm$ 4.7 & 76.4 $\pm$ 3.5  & +2.4 \\
+ Loss reweighting & Gemini + \gls*{wce} + \gls*{cd} & Swin-T & 92.7 $\pm$ 0.9 & 75.7 $\pm$ 3.0 & 75.0 $\pm$ 2.7 & +0.8 \\
+ Scheduler & Gemini + \gls*{wce} + \gls*{lwcd} & DINOv3-Large & 93.2 $\pm$ 0.8 & 76.7 $\pm$ 4.3 & 75.3 $\pm$ 3.0 & +1.0 \\
+ Color-safe aug. & + augmentation & DINOv3-Large & 92.9 $\pm$ 0.5 & 77.3 $\pm$ 4.0 & 74.7 $\pm$ 2.6  & +0.6 \\
+ Segmented bg. & + background blur & DINOv3-Large & 93.0 $\pm$ 0.8 & 78.1 $\pm$ 4.6 & 76.0 $\pm$ 3.0 & +0.8 \\
+ Ensemble & hard voting & Ensemble & \textbf{94.6} $\pm$ 0.4 & \textbf{79.7} $\pm$ 4.0 & \textbf{78.8} $\pm$ 3.1 & \textbf{+1.6} \\
\bottomrule
\end{tabular}
}
\end{table}

The largest single-model gain comes from Gemini-augmented data, which raises macro accuracy from 72.5\% to 74.9\%.
Loss reweighting and \gls*{lwcd} then increase the result to 76.7\%, showing that synthetic data and imbalance-aware optimization are complementary.
Color-safe augmentation and foreground-aware preprocessing provide smaller gains, leading to the best single model at 78.1\% macro accuracy.
Finally, hard voting among the three strongest DINOv3-Large variants yields 79.7\% macro accuracy and 78.8\% macro~F1.

The selected ensemble combines three DINOv3-Large variants trained with Gemini data, WCE, and LWCD: one without augmentation, one with color-safe augmentation, and one with both color-safe augmentation and segmented-background training.
We also evaluated more heterogeneous ensembles combining different architectures, such as Swin-T and ViT-B/16, but these alternatives led to lower macro accuracy.
Therefore, the final composition is deliberately conservative: its members share the same backbone and synthetic source, but differ in visual invariance and foreground~focus.

\subsection{Error Analysis}

To better understand the remaining failures, we manually analyzed all errors produced by the final hard-voting ensemble on three representative folds, selected according to the worst, median, and best macro accuracy.
Each error was independently labeled by two annotators as a \textit{model error}, when the correct color could be inferred from the image, or an \textit{inherently ambiguous} case, when visual evidence was insufficient for reliable color assignment.
Across these folds, the ensemble produced 785 errors.
Considering only annotator consensus labels, 459 errors~(58.5\%) were considered ambiguous and 326~(41.5\%) were considered model errors, with observed agreement~$P_o=73.63\%$ and Cohen's~$\kappa=0.35$.

These categories help contextualize the remaining gap after the proposed approach reaches nearly 80\% macro accuracy in a highly long-tailed surveillance setting.
Model errors suggest room for improvement, whereas ambiguous errors expose a more fundamental limitation of surveillance VCR.
Such cases include infrared nighttime capture, severe underexposure, reflections, occlusion, and borderline hues such as gray~$\leftrightarrow$~silver, silver~$\leftrightarrow$~white, or beige~$\leftrightarrow$~brown.
Remarkably, under this criterion, the estimated upper bound is about 96.8\% micro accuracy and 88.5\% macro accuracy, indicating that the proposed result already closes a substantial part of the achievable gap and that many remaining failures are tied to the visual limits of the~task.

\begin{figure}[!htb]
    \centering
    \captionsetup[subfigure]{
        font=scriptsize,
        captionskip=1pt,
        labelformat=empty,
        justification=raggedright,
        singlelinecheck=false
    }

    \resizebox{0.975\linewidth}{!}{
    \subfloat[{\begin{tabular}[t]{@{}l@{}}~Pred: Black\\~GT: Brown\end{tabular}}]{
        \includegraphics[height=14.2ex]{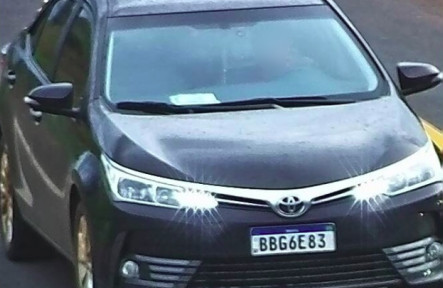}
    } \,
    \subfloat[{\begin{tabular}[t]{@{}l@{}}~Pred: Gray\\~GT: Silver\end{tabular}}]{
        \includegraphics[height=14.2ex]{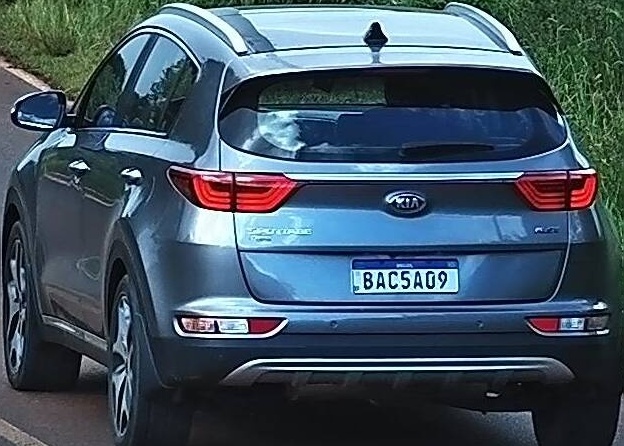}
    } \,
    \subfloat[{\begin{tabular}[t]{@{}l@{}}~Pred: Red\\~GT: Orange\end{tabular}}]{
        \includegraphics[height=14.2ex]{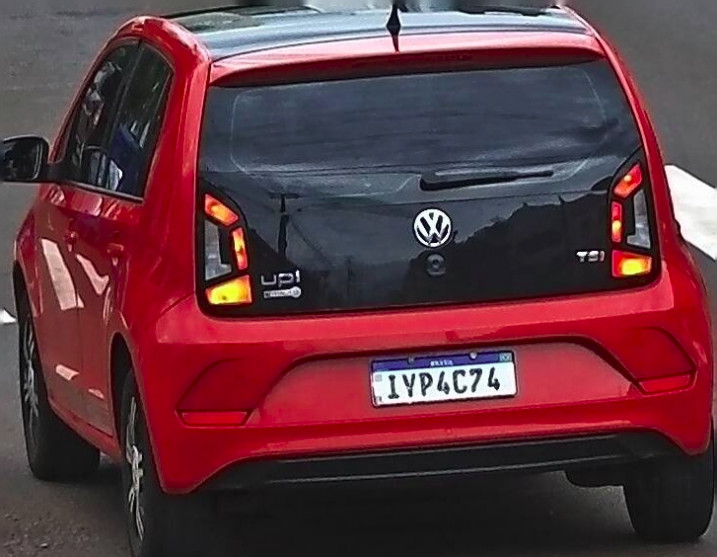}
    } \,
    \subfloat[{\begin{tabular}[t]{@{}l@{}}~Pred: Multicolored\\~GT: White\end{tabular}}]{
        \includegraphics[height=14.2ex]{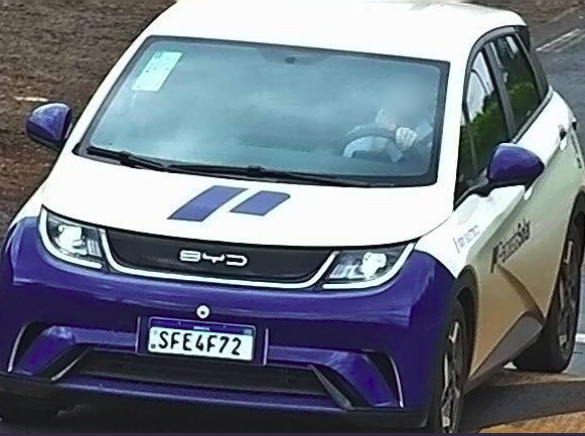}
    }
    }

    \vspace{2.25mm}

    \resizebox{0.975\linewidth}{!}{
    \subfloat[{\begin{tabular}[t]{@{}l@{}}~Pred: Multicolored\\~GT: White\end{tabular}}]{
        \includegraphics[height=13.6ex]{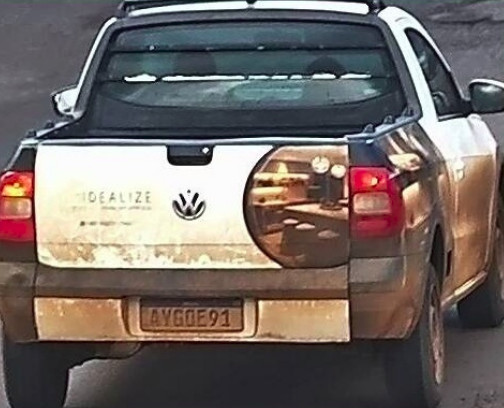}
    } \,
    \subfloat[{\begin{tabular}[t]{@{}l@{}}~Pred: White\\~GT: Silver\end{tabular}}]{
        \includegraphics[height=13.6ex]{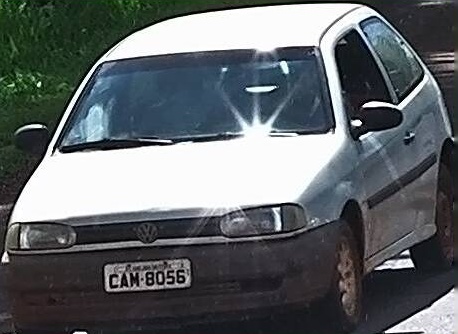}
    } \,
    \subfloat[{\begin{tabular}[t]{@{}l@{}}~Pred: Multicolored\\~GT: Unknown\end{tabular}}]{
        \includegraphics[height=13.6ex]{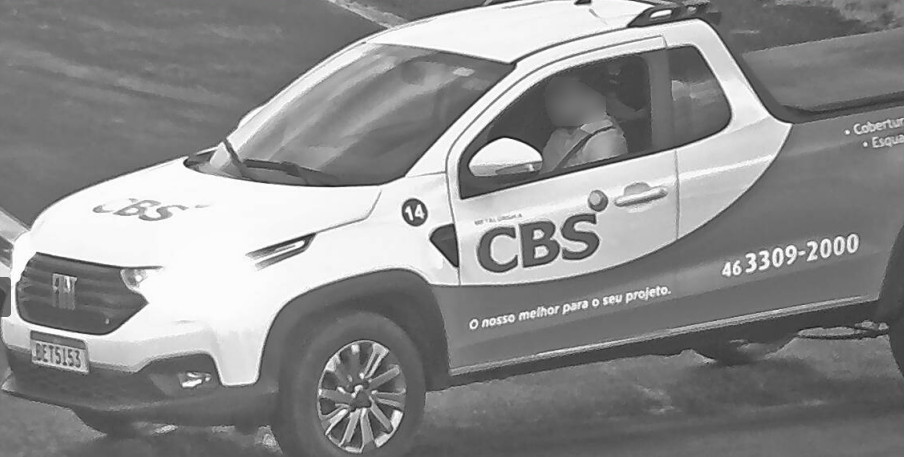}
    } \,
     \subfloat[{\begin{tabular}[t]{@{}l@{}}~Pred: Brown\\~GT: Green\end{tabular}}]{
        \includegraphics[height=13.6ex]{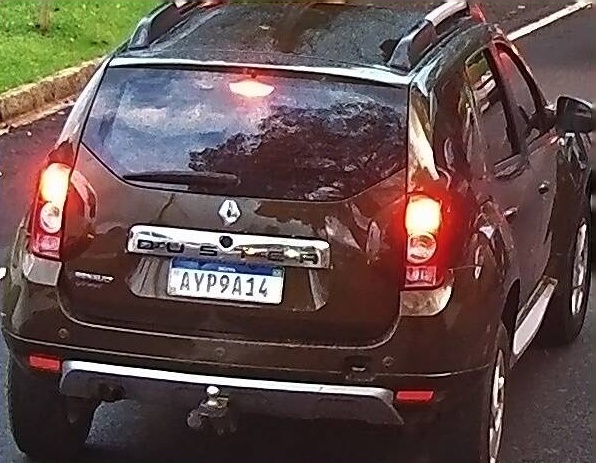}
    }
    }

    \vspace{-1.5mm}
    \caption{Examples of ambiguous misclassifications. In several cases, the ground-truth label~(GT) is difficult to confirm visually because of low illumination, reflections, partial views, or color evidence restricted to a small vehicle region.}
    \label{fig:error_examples}
\end{figure}

Correctable model errors are concentrated in gray~$\rightarrow$~black and silver~$\rightarrow$~white predictions, often associated with low contrast, shadows, and highlights.
Ambiguous errors are dominated by multicolored~$\rightarrow$~white and gray~$\leftrightarrow$~silver cases.
Importantly, the ground-truth labels~(except for the unknown class, which was assigned by the authors of~\cite{lima2026toward} to samples whose color could not be reliably determined) come from official vehicle documentation rather than from visual annotation~\cite{lima2026toward}, so these cases should not be interpreted as annotation errors.
Instead, they often reveal a mismatch between the documented color and the color evidence available in the surveillance image.
For instance, multicolored~$\rightarrow$~white errors commonly occur when only a small vehicle region supports the documented multicolored label, while gray~$\leftrightarrow$~silver cases reflect a weak visual boundary between neutral colors under changing illumination.
\cref{fig:error_examples} illustrates these ambiguities, showing that real-world \gls*{vcr} is limited not only by model capacity but also by the reliability of color evidence in the image.

\section{Conclusions}
\label{sec:conclusions}

This paper presented a comprehensive study of vehicle color recognition under severe class imbalance in real-world surveillance imagery. 
Using UFPR-VeSV~\cite{lima2026toward} as a challenging benchmark, we evaluated synthetic minority-class augmentation together with modern visual representations, imbalance-aware training, learning-rate scheduling, color-safe augmentation, foreground-aware preprocessing, and ensemble fusion. 
The results show that synthetic data is most effective when it is combined with an appropriate training objective: while standard cross-entropy yielded only limited gains, the addition of curated synthetic images made weighted cross-entropy substantially more stable and effective.

Among the evaluated synthetic sources, Gemini 2.0 Flash yielded the greatest performance gains despite contributing fewer images than RunDiffusion/Juggernaut-XL, suggesting that domain alignment and preservation of the surveillance context matter more than synthetic data volume alone.
By combining Gemini-augmented training, DINOv3-Large representations, weighted loss, linear warmup with cosine decay, color-safe augmentation, foreground-aware preprocessing, and hard-voting ensemble fusion, the proposed system achieves a macro accuracy of 79.70\%, outperforming the previous state-of-the-art by 8.20 percentage points~\cite{lima2026toward}.

The manual error analysis further highlights an important limitation of real-world \gls*{vcr}: a large portion of the remaining mistakes are not simply model failures, but visually ambiguous cases that are difficult even for human annotators. 
Such errors are often caused by low illumination, infrared nighttime capture, reflections, occlusions, and weak boundaries between visually similar colors such as gray, silver, white, beige, and brown. 
Therefore, although the proposed pipeline substantially improves class-balanced recognition, the results also show that the practical limits of \gls*{vcr} depend not only on model capacity, but also on the amount and reliability of color evidence available in the image.

Future work will investigate synthetic augmentation in multi-task settings that jointly exploit color, type, make, model, and other vehicle attributes. 
We also plan to explore uncertainty-aware predictions and human-in-the-loop protocols, which may be especially useful when the visual evidence is insufficient for assigning a single reliable color label.

\vspace{-1mm}
\subsubsection{Ethics and Privacy Considerations}

The synthetic images used in this study were generated from textual prompts or by editing images from \dataset~\cite{lima2026toward}. Gemini-based samples may preserve scene elements, including visible license plates and people. Therefore, these images will be distributed and used under the same access, distribution, and research-purpose restrictions applied to the original dataset. We have already obtained authorization from the authors of \dataset to distribute and use the synthetic images under these~conditions.

\iffinal
    \vspace{-1mm}
    \subsubsection{Acknowledgments}

    This study was supported in part by the \textit{Coordenação de Aperfeiçoamento de Pessoal de Nível Superior} (CAPES) -- Finance Code 001, through the \textit{Programa de Excelência Acadêmica} (PROEX); in part by the \textit{Conselho Nacional de Desenvolvimento Científico e Tecnológico} (CNPq) under grant \#~315409/2023-1; and in part by the \textit{Fundação Araucária} under grant \#~078/2026. The authors also thank the \textit{Pontifícia Universidade Católica do Paraná} for the financial support that enabled conference participation.
\else
    \subsubsection{Acknowledgments} 
    
    The acknowledgments are hidden for review.
\fi

\bibliographystyle{splncs04}
\bibliography{bibtex-short-no-doi}

\end{document}